\documentclass{article} %
\usepackage{iclr2027_conference,times}
\usepackage{algorithm}
\usepackage{algpseudocode}

\usepackage{amsmath,amsfonts,bm}

\def\eqref#1{equation~\ref{#1}}

\def\1{\bm{1}}

\DeclareMathAlphabet{\mathsfit}{\encodingdefault}{\sfdefault}{m}{sl}
\SetMathAlphabet{\mathsfit}{bold}{\encodingdefault}{\sfdefault}{bx}{n}

\usepackage{hyperref}
\usepackage{url}
\usepackage{float} 
\usepackage{booktabs}
\usepackage{graphicx}
\usepackage{amsmath}
\usepackage{multirow}
\usepackage{threeparttable}
\usepackage{tabularx}
\usepackage{longtable}
\usepackage{stfloats}
\usepackage{amssymb}
\usepackage{cleveref}
\usepackage{booktabs}
\usepackage{multirow}
\usepackage{booktabs}
\usepackage{multirow}
\usepackage{graphicx}
\usepackage[table]{xcolor}
\RequirePackage{fix-cm}
\usepackage{marvosym}

\usepackage{caption}

\title{FlowState: Execution State as Memory \\ for Long-Horizon LLM Agents}

\iclrfinalcopy

\author{%
  \parbox[t]{\dimexpr\textwidth-2\tabcolsep\relax}{%
    \raggedright\normalfont
    \vspace{-12pt}%
    {\fontsize{11}{13}\selectfont
      \setlength{\tabcolsep}{14pt}%
      \renewcommand{\arraystretch}{1}%
      \begin{tabular}{@{}llll@{}}
        \textbf{Minghao Li}\textsuperscript{*} &
        \textbf{Bangyan Li}\textsuperscript{*} &
        \textbf{Zifan Wang} &
        \textbf{Yulong Li}
        \\[2pt]
        \textbf{Hu Xu} &
        \textbf{Gan Zhang} &
        \textbf{Jingtong Wu} &
        \textbf{Wenqiang Xu}\textsuperscript{†}
      \end{tabular}%
    }\\[5pt]
    {\normalfont\fontsize{10}{12}\selectfont
      Ant International, Ant Group%
    }\\[3pt]
    {\normalfont\ttfamily\fontsize{9}{11}\selectfont
      \mbox{\{lmh499216, libangyan.lby, yugong.xwq\}@ant-intl.com}%
    }\\[4pt]
    {\normalfont\fontsize{9}{11}\selectfont
      \textsuperscript{*} Equal contribution
      \qquad
      \textsuperscript{†} Corresponding author%
    }%
  }%
}

\definecolor{groupblue}{RGB}{235,243,255}
\definecolor{deltagray}{RGB}{235,235,235}
\definecolor{deltagreen}{RGB}{222,240,228}
\definecolor{deltatext}{RGB}{0,112,60}

\newcommand{\perfgain}[1]{%
  {\color{deltatext}\normalfont #1}%
}
\newcommand{\tokensavingbold}[1]{%
  {\color{deltatext}\bfseries #1\%}%
}

\definecolor{deltared}{RGB}{175,65,65}

\begin{document}
\maketitle
\fancyhead{}
\fancyhead[L]{\normalfont Preprint}
\renewcommand{\headrulewidth}{0.4pt}

\begin{abstract}
Long-horizon tasks require LLM agents to continually draw on information from earlier interactions. However, retaining the full history increases context costs, while compressing it risks losing details needed later, and the relevance of historical information often becomes apparent as the task progresses. To address these challenges, we propose \textbf{FlowState}, which treats execution state as memory that can be retained and revisited across requests, unifying current decision-making with the reuse of historical information. FlowState preserves semantically typed state nodes, their relations, and references to raw tool observations, separating persistent retention from on-demand access. Within a single execution loop, Incremental State Update (\textbf{ISU}) maintains the current state based on new inputs and feedback, while Progressive State Access (\textbf{PSA}) progressively reveals historical states and supporting evidence as needed during reasoning. Together, these mechanisms enable agents to reassess prior decisions in light of new information and guide subsequent actions. Compared with a full-context baseline using the same DeepSeek-V4-Flash model, FlowState improves the average success rate on MemoryArena and the average pass rate on $\tau^3$-Bench by \textbf{4.55} and \textbf{13.95} percentage points, respectively, while reducing total token consumption by \textbf{43.2\%} and \textbf{40.6\%}. These results demonstrate the performance and efficiency advantages of FlowState on long-horizon tasks.
\end{abstract}

\section{Introduction}
\label{sec:intro}

\begin{figure*}[t]
  \centering
  \includegraphics[width=0.95\textwidth]{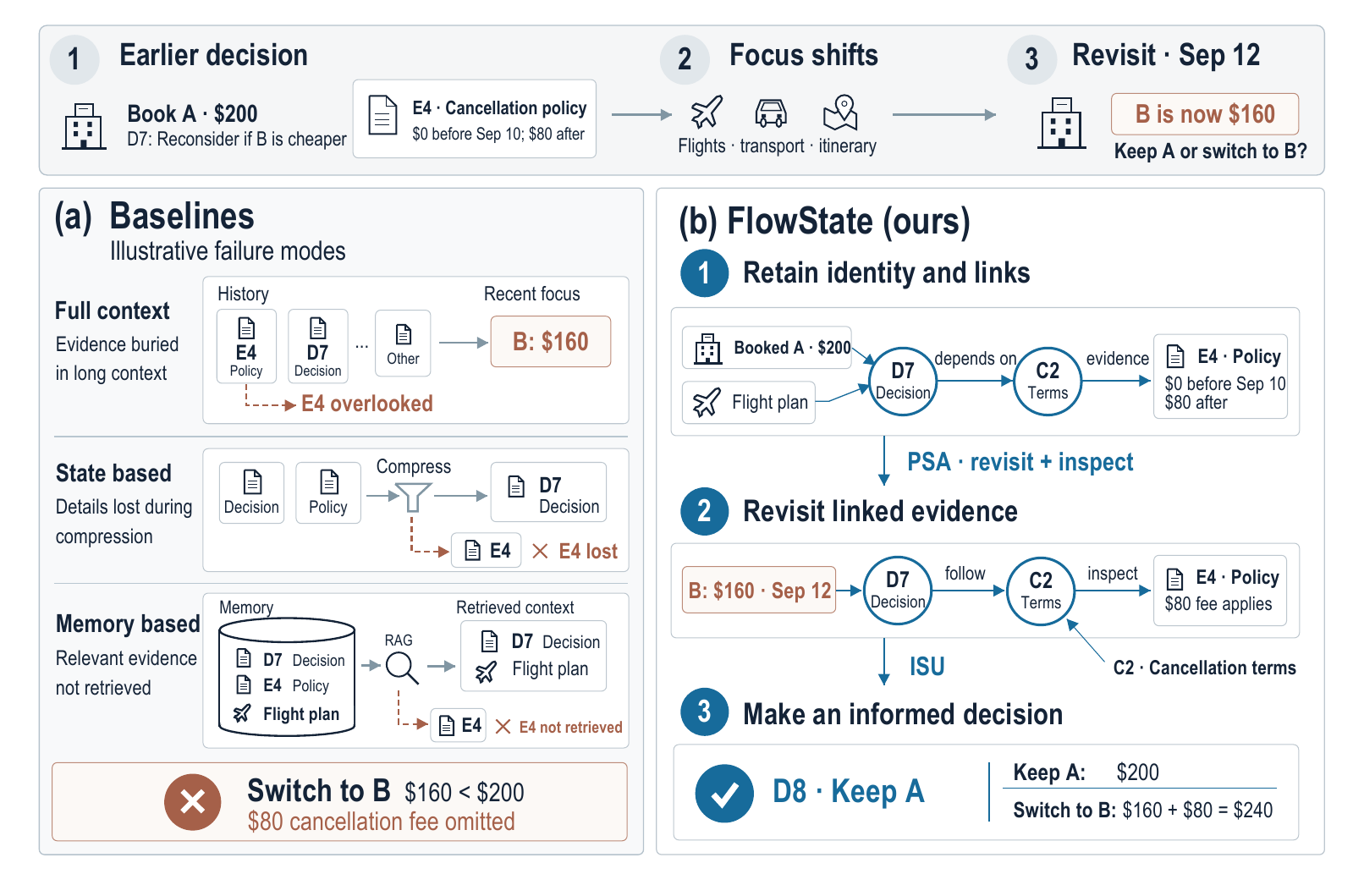}
\caption{
\textbf{Revisiting earlier decisions through linked evidence.}
A user books hotel A for \$200, planning to reconsider if B becomes cheaper, then handles other travel arrangements.
On September 12, B drops to \$160, but A's \$80 cancellation fee makes switching costlier (\$240 vs.\ \$200).
\textbf{(a)} Illustrative baseline failures: the cancellation policy may be overlooked in full context, lost during compression, or stored but not retrieved.
\textbf{(b)} FlowState preserves D7 and its links to cancellation terms C2 and policy evidence E4 beyond the current view.
PSA revisits D7 and linked evidence; ISU combines the new price with this evidence to form an updated decision to keep A (D8).
}
\label{fig:motivation}
\end{figure*}

As large language models improve in reasoning, tool use, and environmental interaction, LLM-based agents can complete complex tasks through multiple steps of action and observation~\citep{yao2022react}. Long-horizon success, however, requires agents to track execution progress and use accumulated information to guide subsequent actions~\citep{he2026memoryarena}. Travel arrangements must respect existing constraints, while mathematical derivations may depend on earlier intermediate results. Such dependencies persist across subtasks and requests, making it essential for information acquired during execution to remain available for future decisions.

Maintaining this information poses a trade-off. Full interaction histories increase context costs and can obscure relevant details~\citep{singh2026agent}, while compressing these histories may discard information needed for subsequent actions~\citep{kang2025acon}. State-maintenance methods organize facts, constraints, and progress into working representations~\citep{rozanov2025stateact,uddin2026ledgeragent,guo2026state}. ZipAct, for example, replaces the growing history with a continuously updated compact state~\citep{panzipact}. Yet repeated compression may discard details needed later~\citep{ye2025agentfold}, whereas retaining historical information in the working state can clutter the context. Agents therefore need to separate persistent information retention from current use.

Long-term memory methods preserve information through external storage and retrieval~\citep{packer2023memgpt,chhikara2025mem0,xu2026mem,rasmussen2025zep}, but when and what to retrieve often emerge only during reasoning~\citep{ji2026memory}. New feedback or changing constraints can make earlier information relevant again, and a retrieved record may require further examination of related states or original evidence. In Figure~\ref{fig:motivation}, a user books hotel A and later learns that hotel B's price has dropped. Whether to switch depends on both the new price and the earlier booking's cancellation terms. Memory access must therefore unfold alongside decision-making, enabling agents to retrieve and combine historical information as needs emerge.

Recent work uses current state to select historical experiences~\citep{wang2026samem} and intermediate evidence to adjust memory access paths~\citep{ji2026memory}. This motivates our central question: \textbf{Can execution state itself serve as a unified representation connecting current decisions with long-term memory?} Retaining execution states across requests and restoring them with related information and original evidence would let agents build on prior work without repeatedly processing the full history. Execution state could thus serve as both a working representation and a reusable memory unit.

We propose FlowState, an agent memory framework centered on persistent execution state. FlowState represents user requirements, environmental information, agent judgements, and execution artifacts as semantically typed nodes, preserving their relations and references to raw tool observations. Full contents persist in a Persistent State Repository, while identifiers and concise summaries form a Historical State Index included in the model context. The agent selectively expands historical states through the same representation used for current states, separating persistent retention from context visibility to keep the working context compact.

FlowState integrates two complementary mechanisms into a single execution loop. Incremental State Update (ISU) creates and revises current states from new inputs and execution feedback without reconstructing the entire state. Progressive State Access (PSA) retrieves historical content through the index or relations in already disclosed states, tracing back to raw observations as needed. Retrieved information informs subsequent actions and state updates. At the end of each request, the resulting states and updated index are retained for future reuse.

We evaluate FlowState on MemoryArena~\citep{he2026memoryarena} and $\tau^3$-Bench~\citep{barres2025tau}, covering shopping, travel planning, formal reasoning, and service interactions governed by business rules. Compared with the full-context baseline using the same DeepSeek-V4-Flash model, FlowState improves the average success rate on MemoryArena and the average pass rate on $\tau^3$-Bench by 4.55 and 13.95 percentage points, respectively, while reducing total token consumption by 43.2\% and 40.6\%. Further analyses show that ISU and PSA play complementary roles, with strong performance on later subtasks and sustained performance and efficiency gains in evaluated settings with strong baseline models.

Our main contributions are as follows:
\begin{itemize}
    \item \textbf{A unified memory representation centered on execution state.}
    We extend execution states into memory units that can be reused across requests, connecting decisions with historical information through semantic types, node relations, and evidence references.

    \item \textbf{Incremental maintenance and progressive access within a single execution loop.}
    ISU and PSA maintain states for the current request and access historical information on demand, allowing memory to accumulate during execution and inform decisions.

    \item \textbf{Improved success rates and token efficiency across diverse long-horizon tasks.}
    FlowState improves task success rates while reducing token consumption on MemoryArena and $\tau^3$-Bench. We further examine its effectiveness through ablations, analyses across subtask depths, and experiments across base models.
\end{itemize}

\section{FlowState}
\label{sec:method}

We consider a sequential interaction task consisting of user requests $q_1,\ldots,q_K$. For request $q_k$, the agent interacts with the environment over $T_k$ ReAct decision steps, each corresponding to one model invocation. A fixed policy $\pi_\theta$ makes decisions based on the information available at each step; its parameters remain unchanged throughout the interaction. Constraints, observations, action outcomes, and assessments from earlier requests may become relevant again, while the specific historical information needed often emerges only as the current request unfolds. FlowState therefore persists execution state across requests and provides access to it on demand, allowing historical states to inform subsequent decisions.

\subsection{Persistent Execution State}
\label{sec:persistent-execution-state}

FlowState's persistent execution state comprises a runtime-maintained \emph{Persistent State Repository}, a \emph{Historical State Index} for the current request, and an \emph{Active State} that evolves during execution.

\paragraph{Persistent State Repository ($\mathcal{M}_k$).}

We define $\mathcal{M}_k$ as the persistent state repository at the start of request $q_k$. It stores prior user requests, state nodes, and relations among nodes. Each node records its source request, semantic identifier, category, and full content. A raw tool result is also retained as an accessible state node if it is referenced by a state newly created by the model. The runtime maintains $\mathcal{M}_k$, but does not place it directly in the model context. At the end of $q_k$, newly created states and relations retained for subsequent requests are incorporated into $\mathcal{M}_{k+1}$.

\paragraph{Historical State Index ($G_k$).}

We define $G_k=\operatorname{Index}(\mathcal{M}_k)$ as a lightweight index of historical states constructed at the start of request $q_k$.

The index covers all preceding requests $q_1,\ldots,q_{k-1}$. For each request, a state entry contains only its semantic identifier and a concise summary maintained for that state. The lightweight index $G_k$ is included in the model context for request $q_k$.

\paragraph{Active State ($S_{k,t}$).}

We define $S_{k,t}$ as the Active State available for decision-making at step $t$ of request $q_k$. It contains states created during the current request, together with disclosed historical states and their relations. The full contents of undisclosed historical states are excluded from $S_{k,t}$.

Nodes in the Active State follow a taxonomy of five state types. FlowState classifies information by source and role: user-side \emph{preferences} record the user's preferences, requirements, and constraints; environment-side \emph{knowledge} records tool observations, while \emph{attributes} capture entity-level information distilled from those observations; and agent-side \emph{judgements} record the agent's assessments of task objectives, plans, decisions, and risks, while \emph{artifacts} record reusable outputs produced by the agent. Details on the semantic scope of each state type are provided in Appendix~\ref{app:state-types}.

\begin{figure}[t]
\centering
\includegraphics[width=0.95\linewidth]{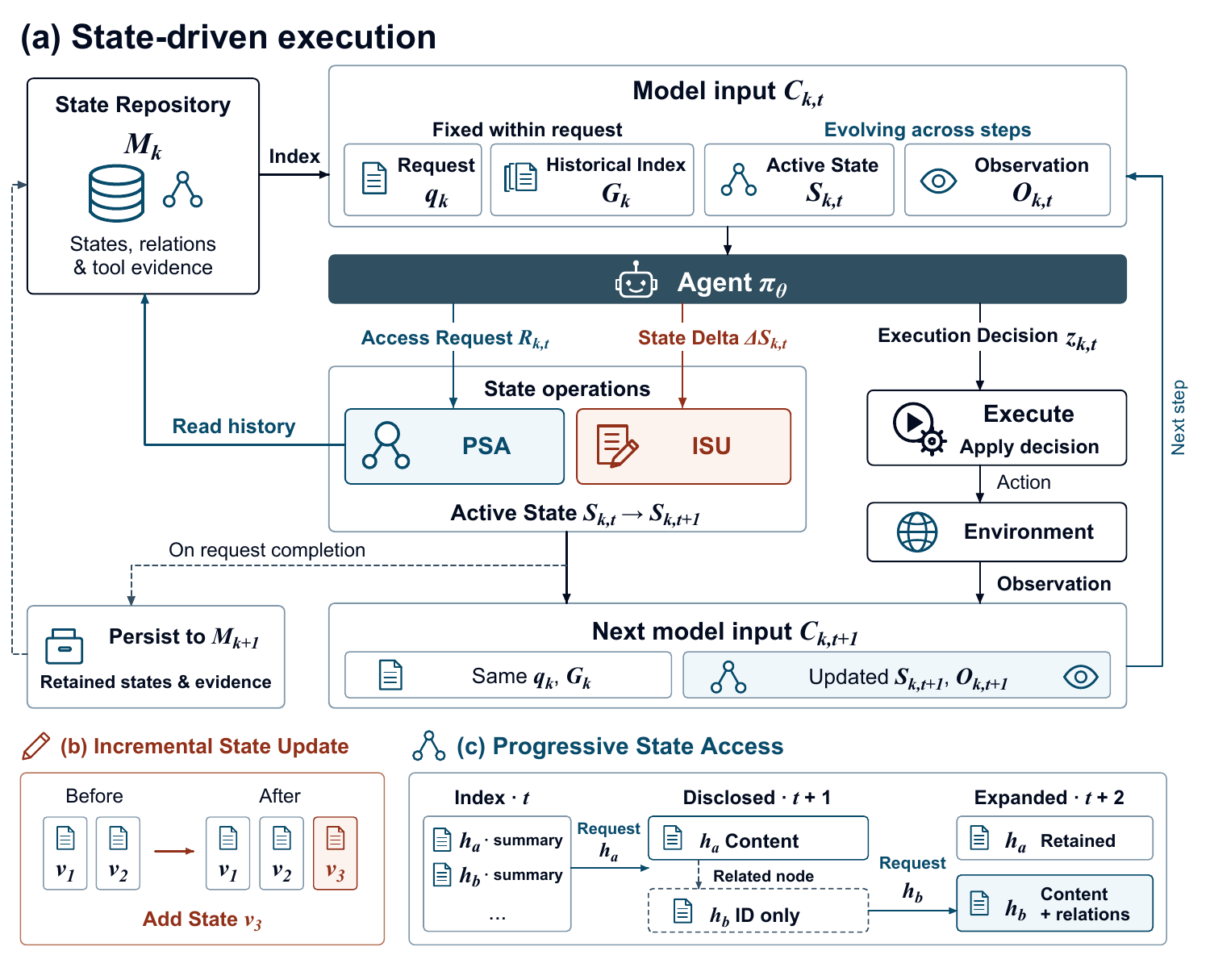}
\caption{\textbf{FlowState execution loop and state operations.}
(a) Each model invocation proposes a state delta, a historical-state access request, and an execution decision. ISU and PSA update the Active State; retained states and evidence are persisted at request completion.
(b) ISU is illustrated by adding $v_3$ to the current-request state.
(c) PSA discloses $h_a$, whose revealed relation guides a later access to $h_b$.}
\label{fig:flowstate-framework}
\end{figure}

\subsection{State Operations}
\label{sec:state-operations}

FlowState maintains $S_{k,t}$ through two complementary operations: \emph{Incremental State Update} (ISU) commits local state changes in response to new information, while \emph{Progressive State Access} (PSA) progressively accesses historical states as information needs arise during execution. The model proposes both types of requests from the same context. After validation against $S_{k,t}$ under execution rules $P$ (Appendix~\ref{app:execution-rules}), their effects are combined to produce $S_{k,t+1}$. Concurrent work, MAGE~\citep{chen2026beyond}, organizes execution
history as a hierarchical state tree with active-path context
and branching revision.
FlowState instead uses semantically typed state nodes that are
individually addressable across requests, separating persistent
retention from visibility in the Active State.
Through PSA and ISU, historical states and supporting evidence
are progressively disclosed to inform current-request state updates.

\subsubsection{Incremental State Update (ISU)}

At step $t$, the model uses the current context $C_{k,t}$ to produce a candidate state delta $\Delta S_{k,t}$, comprising three operations on state nodes: $\mathsf{Add}$, $\mathsf{Update}$, and $\mathsf{Remove}$. $\mathsf{Add}$ creates a state for the current request; $\mathsf{Update}$ and $\mathsf{Remove}$ apply only to states created during the current request. The state delta is validated before it is committed:

\begin{equation}
\widehat{\Delta S}_{k,t}
=\operatorname{Validate}_{\mathrm{upd}}(\Delta S_{k,t}\mid P,S_{k,t}),
\qquad
S_{k,t}^{+}
=\operatorname{Commit}(S_{k,t},\widehat{\Delta S}_{k,t}).
\label{eq:isu-update}
\end{equation}

The validated state delta $\widehat{\Delta S}_{k,t}$ is committed to $S_{k,t}$, yielding the intermediate Active State $S_{k,t}^{+}$ after ISU. The model need not regenerate the entire state at every step. For example, a price observation returned by the environment may update a \emph{judgement} about lodging selection created during the current request, while an unchanged budget \emph{preference} remains intact.

\subsubsection{Progressive State Access (PSA)}

PSA discloses historical state information on demand as information needs arise during execution. At step $t$, the model uses state identifiers and their summaries in $G_k$, together with disclosed relations in $S_{k,t}$, to specify a set $R_{k,t}$ of historical states to access; if no historical states are needed, $R_{k,t}=\varnothing$. After the access request is validated against the execution rules, the corresponding states are disclosed from $\mathcal{M}_k$:

\begin{equation}
\widehat{R}_{k,t}
=\operatorname{Validate}_{\mathrm{acc}}(R_{k,t}\mid P,G_k,S_{k,t}),\quad
S_{k,t+1}
=S_{k,t}^{+}\cup\operatorname{Disclose}(\mathcal{M}_k,S_{k,t},\widehat{R}_{k,t}).
\label{eq:psa-transition}
\end{equation}

$\operatorname{Disclose}$ returns the contents and relations of validated target states $\widehat{R}_{k,t}$ that are not already in $S_{k,t}$; these are added to $S_{k,t}^{+}$ to obtain $S_{k,t+1}$. Each relation specifies its type and the identifier of the related node, allowing the model to decide whether to access that node in a subsequent step. Disclosing a relation does not reveal the full contents of the related node.

\begin{algorithm}[!t]
\caption{FlowState State-Driven Execution Loop.}
\label{alg:flowstate}
\begin{algorithmic}[1]
\Require Requests $q_1,\ldots,q_K$, fixed policy $\pi_\theta$, execution rules $P$, and environment tools
\State Initialize Persistent State Repository $\mathcal{M}_1$
\For{$k=1,\ldots,K$}
    \State $G_k\leftarrow\operatorname{Index}(\mathcal{M}_k)$
    \State Initialize $S_{k,1}\leftarrow\varnothing$ and $O_{k,1}\leftarrow\varnothing$; set $t\leftarrow 1$
    \While{$q_k$ is unfinished}
        \State $C_{k,t}\leftarrow\operatorname{Serialize}(q_k,O_{k,t},G_k,S_{k,t})$
        \State Sample $y_{k,t}\sim\pi_\theta(\cdot\mid C_{k,t})$ subject to $y_{k,t}\neq(\varnothing,\varnothing,\varnothing)$
        \State $(\Delta S_{k,t},R_{k,t},z_{k,t})\leftarrow y_{k,t}$
        \State $\widehat{\Delta S}_{k,t}\leftarrow\operatorname{Validate}_{\mathrm{upd}}(\Delta S_{k,t}\mid P,S_{k,t})$
        \State $\widehat{R}_{k,t}\leftarrow\operatorname{Validate}_{\mathrm{acc}}(R_{k,t}\mid P,G_k,S_{k,t})$
        \State $S_{k,t}^{+}\leftarrow\operatorname{Commit}(S_{k,t},\widehat{\Delta S}_{k,t})$
        \State $S_{k,t+1}\leftarrow S_{k,t}^{+}\cup\operatorname{Disclose}(\mathcal{M}_k,S_{k,t},\widehat{R}_{k,t})$
        \State $O_{k,t+1}\leftarrow\varnothing$
        \If{$z_{k,t}$ is a final response}
            \State Output $z_{k,t}$ and \textbf{break}
        \ElsIf{$z_{k,t}$ is an environment action}
            \State Execute $z_{k,t}$ and receive tool observation $O_{k,t+1}$
        \EndIf
        \State $t\leftarrow t+1$
    \EndWhile
    \State $\mathcal{M}_{k+1}\leftarrow\operatorname{Persist}(\mathcal{M}_k,q_k,S_{k,T_k+1},\mathbf{O}_k)$
\EndFor
\end{algorithmic}
\end{algorithm}

\subsection{State-Driven Execution}
\label{sec:state-driven-execution}

Figure~\ref{fig:flowstate-framework} situates the ISU and PSA operations described above within the state-driven execution loop.

At step $t$ of request $q_k$, FlowState serializes the current request, the most recent tool observation, the Historical State Index, and the Active State into a model input. Let $y_{k,t}=(\Delta S_{k,t},R_{k,t},z_{k,t})$ denote the model output, whose components are the candidate state delta, the target historical states to access, and the execution decision, respectively:

\begin{equation}
C_{k,t}=\operatorname{Serialize}(q_k,O_{k,t},G_k,S_{k,t}),\quad
y_{k,t}\sim\pi_\theta(\cdot\mid C_{k,t}),
\quad\text{s.t. }y_{k,t}\neq(\varnothing,\varnothing,\varnothing).
\label{eq:flowstate-decision}
\end{equation}

Here $C_{k,t}$ is the model input at this step; $O_{k,t}$ is the tool observation produced by the preceding step, or $\varnothing$ if no tool was called; and $z_{k,t}$ is an environment action, a final response, or $\varnothing$. Each of the three output components may be empty individually, but they cannot all be empty simultaneously.

After the model produces its output, ISU and PSA process their respective requests against $S_{k,t}$ and combine their effects to obtain $S_{k,t+1}$. The execution decision $z_{k,t}$ then determines whether to perform an environment action, terminate the current request, or proceed to the next model invocation. Information disclosed by PSA at step $t$ can inform decisions only from the next model invocation onward. For actions that depend on this information, the model can set $z_{k,t}=\varnothing$ and defer the decision to the next step using $S_{k,t+1}$. An empty state delta results in no ISU update, while an empty access request results in no PSA disclosure. Algorithm~\ref{alg:flowstate} gives the complete procedure.

Let $\mathbf{O}_k=(O_{k,1},\ldots,O_{k,T_k+1})$ denote the sequence of tool observations for request $q_k$. Upon completion of the request, the Persistent State Repository is updated:

\begin{equation}
\mathcal{M}_{k+1}
=\operatorname{Persist}(\mathcal{M}_k,q_k,S_{k,T_k+1},\mathbf{O}_k),
\qquad
G_{k+1}=\operatorname{Index}(\mathcal{M}_{k+1}).
\label{eq:cross-request-persistence}
\end{equation}

$\operatorname{Persist}$ incorporates request $q_k$ and the newly created states and relations retained at the end of the request into $\mathcal{M}_{k+1}$, and retains any raw tool results referenced by those states as \emph{knowledge} nodes. New states record $q_k$ as their source request; historical states retain their original source requests and cannot be deleted. $\operatorname{Index}$ then generates $G_{k+1}$ from $\mathcal{M}_{k+1}$, listing state identifiers and their summaries through request $q_k$, grouped by request, to guide subsequent PSA access.

\section{Experiments}
\label{sec:experiments}

\subsection{Experimental Setup}
\label{sec:experiment-setup}

We evaluate FlowState on a subset of MemoryArena environments (Bundled Web Shopping, Group Travel Planning, and Formal Reasoning: Math and Physics)~\citep{he2026memoryarena} and on $\tau^3$-Bench (Airline and Retail)~\citep{barres2025tau}. The former tests information reuse across interdependent subtasks; the latter tests policy-constrained tool use with simulated users.

We compare FlowState against Full Context (full interaction history), memory-system methods (ReasoningBank~\citep{ouyang2026reasoningbank}, BM25~\citep{robertson2009bm25}, and Mem0), a state-tracking method (ZipAct), and a hybrid method (ZipAct+BM25). All methods use DeepSeek-V4-Flash~\citep{deepseek2026v4} in the primary comparison; Full Context is also evaluated with GPT-5.6-terra~\citep{openai2026gpt56} and Qwen3.5-397B-A17B~\citep{qwenteam2026qwen35}.

MemoryArena reports Success Rate (SR), Progress Score (PS), and soft Progress Score (sPS) for Travel. $\tau^3$-Bench reports Pass Rate, DB Accuracy, and Action Accuracy. We report total prompt and generation tokens (\#Tok). Benchmark, scoring, and implementation details appear in Appendix~\ref{app:experimental-details}.

\subsection{Main Results}
\label{sec:main-results}

\begin{table*}[t]
\centering
\caption{
Results on MemoryArena.
Performance metrics are reported in percent;
\#Tok. denotes total token usage.
Bold and underlined performance scores indicate the best
and second-best results, respectively.
$\Delta$ denotes changes relative to DeepSeek-V4-Flash
with full context (FC): absolute percentage-point changes
for performance and relative percentage changes for token usage
(negative values indicate savings).
}
\label{tab:memoryarena-main}

\small
\renewcommand{\arraystretch}{1.12}
\setlength{\tabcolsep}{4.0pt}

\resizebox{\textwidth}{!}{%
\begin{tabular}{
l
ccc
cccc
ccc
ccc
}
\toprule
&
\multicolumn{3}{c}{\multirow{2}{*}{\textbf{Bundled Web Shopping}}}
&
\multicolumn{4}{c}{\multirow{2}{*}{\textbf{Group Travel Planning}}}
&
\multicolumn{6}{c}{\textbf{Formal Reasoning}}
\\
\cmidrule(lr){9-14}
&
\multicolumn{3}{c}{}
&
\multicolumn{4}{c}{}
&
\multicolumn{3}{c}{\textbf{Math}}
&
\multicolumn{3}{c}{\textbf{Physics}}
\\
\cmidrule(lr){2-4}
\cmidrule(lr){5-8}
\cmidrule(lr){9-11}
\cmidrule(lr){12-14}

\textbf{Method}
& \textbf{SR $\uparrow$}
& \textbf{PS $\uparrow$}
& \textbf{\#Tok. $\downarrow$}
& \textbf{SR $\uparrow$}
& \textbf{PS $\uparrow$}
& \textbf{sPS $\uparrow$}
& \textbf{\#Tok. $\downarrow$}
& \textbf{SR $\uparrow$}
& \textbf{PS $\uparrow$}
& \textbf{\#Tok. $\downarrow$}
& \textbf{SR $\uparrow$}
& \textbf{PS $\uparrow$}
& \textbf{\#Tok. $\downarrow$}
\\
\midrule

\rowcolor{groupblue}
\multicolumn{14}{l}{\textbf{Full Context}}
\\
DeepSeek-V4-Flash
& 0.00 & 21.11 & 52.65M
& \underline{0.37} & 3.09 & \underline{88.43} & 231.59M
& 30.00 & 50.36 & 40.60M
& \underline{55.00} & \underline{72.50} & 3.68M
\\
GPT-5.6-terra
& 0.00 & \underline{33.89} & 313.95M
& -- & -- & -- & --
& \textbf{42.50} & \underline{54.23} & 14.59M
& 50.00 & 62.60 & 2.76M
\\
Qwen3.5-397B-A17B
& 0.00 & 19.56 & 134.82M
& \underline{0.37} & \underline{6.96} & 60.99 & 223.70M
& 30.00 & 35.27 & 19.97M
& 50.00 & 55.56 & 2.02M
\\

\midrule
\rowcolor{groupblue}
\multicolumn{14}{l}{\textbf{Memory System}}
\\
ReasoningBank
& 0.00 & 31.11 & 92.72M
& 0.00 & 3.48 & 48.13 & 49.35M
& 32.50 & 45.63 & 7.90M
& \underline{55.00} & 63.97 & 1.88M
\\
BM25
& \underline{0.67} & 22.11 & 32.98M
& 0.00 & 5.24 & 65.09 & 69.82M
& 22.50 & 49.83 & 10.06M
& 45.00 & 60.63 & 1.27M
\\
Mem0
& 0.00 & 25.78 & 59.36M
& 0.00 & 3.37 & 40.74 & 107.22M
& 25.00 & 41.96 & 6.61M
& 35.00 & 53.57 & 1.33M
\\

\midrule
\rowcolor{groupblue}
\multicolumn{14}{l}{\textbf{State Tracking}}
\\
ZipAct
& 0.00 & 17.78 & 33.98M
& 0.00 & 3.81 & 25.22 & 201.86M
& 30.00 & 50.66 & 12.28M
& \underline{55.00} & 68.54 & 3.33M
\\

\midrule
\rowcolor{groupblue}
\multicolumn{14}{l}{\textbf{State Tracking + Memory System}}
\\
ZipAct+BM25
& 0.00 & 16.11 & 65.92M
& 0.00 & 4.29 & 19.89 & 175.40M
& 30.00 & 53.09 & 9.85M
& 40.00 & 61.18 & 2.21M
\\

\midrule
\textbf{FlowState (Ours)}
& \textbf{5.33} & \textbf{42.89} & 44.83M
& \textbf{0.74} & \textbf{12.57} & \textbf{91.87} & 128.93M
& \underline{37.50} & \textbf{55.53} & 10.96M
& \textbf{60.00} & \textbf{73.96} & 1.78M
\\
\rowcolor{deltagreen}
{\footnotesize $\Delta$ \textit{vs. V4-Flash (FC)}}
& \perfgain{+5.33} & \perfgain{+21.78} & \tokensavingbold{-14.86}
& \perfgain{+0.37} & \perfgain{+9.48} & \perfgain{+3.44}
& \tokensavingbold{-44.33}
& \perfgain{+7.50} & \perfgain{+5.17} & \tokensavingbold{-73.01}
& \perfgain{+5.00} & \perfgain{+1.46} & \tokensavingbold{-51.69}
\\
\bottomrule
\end{tabular}%
}
\end{table*}

\begin{table}[t]
\centering
\caption{
Results on $\tau^3$-Bench using DeepSeek-V4-Flash for both methods.
PR, DB, and Act.\ denote pass rate, database accuracy, and
action accuracy (\%), respectively; \#Tok.\ denotes total token usage.
Bold marks better performance.
$\Delta$ is relative to Full Context: percentage points for
performance and percentage changes for tokens (negative means savings).
}
\label{tab:taubench}

\newcommand{\tauup}[1]{%
  {\footnotesize\color{deltagreen}\bfseries\boldmath #1}%
}
\newcommand{\taudown}[1]{%
  {\footnotesize\color{deltared} #1}%
}

\small
\renewcommand{\arraystretch}{1.2}
\setlength{\tabcolsep}{4pt}

\begin{tabular}{@{}l cccccccc@{}}
\toprule
& \multicolumn{4}{c}{\textbf{Airline}}
& \multicolumn{4}{c}{\textbf{Retail}}
\\
\cmidrule(lr){2-5}
\cmidrule(lr){6-9}
\textbf{Method}
& PR $\uparrow$
& DB $\uparrow$
& Act. $\uparrow$
& \#Tok. $\downarrow$
& PR $\uparrow$
& DB $\uparrow$
& Act. $\uparrow$
& \#Tok. $\downarrow$
\\
\midrule
Full Context
& 58.0 & 70.5 & 81.4 & 21.03M
& 73.7 & 80.7 & \textbf{90.4} & 23.55M
\\
\textbf{FlowState (Ours)}
& \textbf{78.0} & \textbf{78.0} & \textbf{83.8} & 8.16M
& \textbf{81.6} & \textbf{86.0} & 85.5 & 18.32M
\\
\rowcolor{deltagreen}
{\footnotesize $\Delta$ \textit{vs. Full Context}}
& \perfgain{+20.0}
& \perfgain{+7.5}
& \perfgain{+2.4}
& \tokensavingbold{-61.21}
& \perfgain{+7.9}
& \perfgain{+5.3}
& \taudown{$-4.9$}
& \tokensavingbold{-22.24}
\\
\bottomrule
\end{tabular}
\end{table}

\begin{figure*}[t]
    \centering
    \includegraphics[width=0.88\textwidth]
        {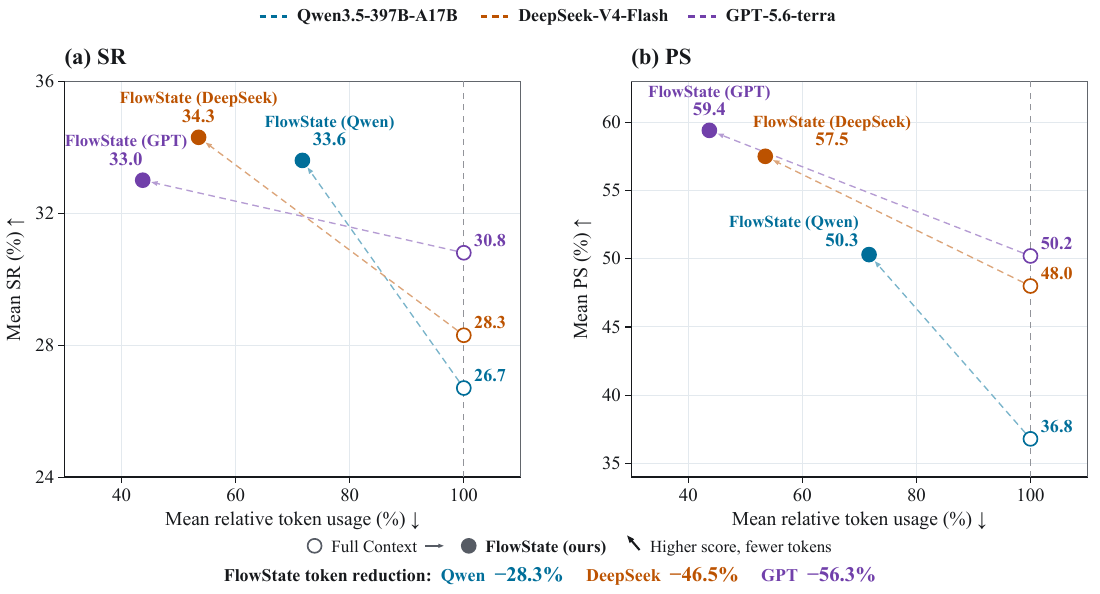}
    \caption{
        \textbf{FlowState consistently improves both performance and token
        efficiency across base models.}
        SR and PS are equally weighted averages over Web Shopping, Math,
        and Physics.
        Relative token usage is normalized to the corresponding
        Full Context baseline within each task and then averaged
        with equal weights across tasks.
        Hollow and filled circles denote Full Context and FlowState,
        respectively; dashed arrows connect results for the same base model.
    }
    \label{fig:flowstate-performance-efficiency}
\end{figure*}

\paragraph{FlowState improves task success and progress on MemoryArena.}
Table~\ref{tab:memoryarena-main} shows FlowState leads on eight of nine task performance metrics. On Bundled Web Shopping, FlowState achieves a PS of 42.89\%, outperforming the strongest memory-system baseline by 11.78 percentage points. On Group Travel Planning, it reaches a PS of 12.57\%, exceeding the best baseline by 5.61 percentage points. FlowState also achieves the best Math PS and the best SR and PS on Physics.

In contrast, directly combining state tracking with historical retrieval does not yield consistent gains: ZipAct+BM25 outperforms both standalone components only on Math PS. FlowState surpasses ZipAct+BM25 on all nine performance metrics, indicating that, on the interdependent multi-session tasks evaluated in MemoryArena, its design of treating execution states as revisitable memory has a consistent performance advantage over this direct combination.

These gains also come with lower context overhead. Across four environments, FlowState reduces total token consumption by 43.2\% compared with Full Context using the same DeepSeek-V4-Flash model, showing it can improve task success and progress while keeping context overhead low.

\paragraph{FlowState's benefits extend beyond MemoryArena.}
To assess generality across benchmarks, we further evaluate FlowState on $\tau^3$-Bench, which differs from MemoryArena in both task format and evaluation criteria. As shown in Table~\ref{tab:taubench}, using the same DeepSeek-V4-Flash model, FlowState improves Pass Rate over Full Context by 20.0 and 7.9 percentage points on Airline and Retail, respectively, and DB Accuracy by 7.5 and 5.3 percentage points, while reducing token consumption by 61.2\% and 22.2\%. Although Action Accuracy on Retail decreases by 4.9 percentage points, task success improves in both domains. These results show that FlowState's benefits extend beyond MemoryArena, providing further evidence of its generality across different types of interactive tasks.

\textbf{Joint Performance and Efficiency Gains Across Models.}
To assess whether FlowState's benefits persist across base models, we evaluate it with DeepSeek-V4-Flash, GPT-5.6-terra, and Qwen3.5-397B-A17B, comparing each with Full Context using the same model. Figure~\ref{fig:flowstate-performance-efficiency} shows higher mean SR and PS for all three models, alongside 28.3\%--56.3\% lower mean relative token usage. These results suggest that FlowState's performance and token-efficiency gains are not specific to a single base model. Per-task results appear in Appendix Table~\ref{tab:flowstate_model_comparison}.

\definecolor{ablationblue}{RGB}{234,242,255}

\begin{table}[t]
\centering
\caption{
Ablation results on Group Travel Planning and Bundled Web Shopping.
Performance metrics are in percent; \#Tok.\ denotes total token usage.
Bold scores mark the best performance within each environment.
}
\label{tab:ablation}
\small
\renewcommand{\arraystretch}{1.18}
\setlength{\tabcolsep}{6pt}

\begin{tabular}{l cccc ccc}
\toprule
& \multicolumn{4}{c}{\textbf{Travel Planning}}
& \multicolumn{3}{c}{\textbf{Web Shopping}}
\\
\cmidrule(lr){2-5}
\cmidrule(lr){6-8}
\textbf{Variant}
& SR $\uparrow$
& PS $\uparrow$
& sPS $\uparrow$
& \#Tok.\ $\downarrow$
& SR $\uparrow$
& PS $\uparrow$
& \#Tok.\ $\downarrow$
\\
\midrule
\rowcolor{ablationblue}
\textbf{FlowState}
& \textbf{0.74}
& \textbf{12.57}
& \textbf{91.87}
& 128.93M
& \textbf{5.33}
& 42.89
& 44.83M
\\
w/o PSA
& 0.37
& 9.36
& 87.18
& 160.27M
& 3.33
& \textbf{44.67}
& 115.37M
\\
w/o ISU
& 0.00
& 9.15
& 90.34
& 93.16M
& 0.00
& 32.11
& 69.01M
\\
\bottomrule
\end{tabular}
\end{table}

\section{Analysis}
\subsection{Component Ablation}

We compare FlowState with two ablation variants to assess the roles of ISU and PSA. As shown in Table~\ref{tab:ablation}, \emph{w/o PSA} removes the Historical State Index and progressive access, instead retaining the full contents of all ISU-generated historical states in the Active State across subtasks. In contrast, \emph{w/o ISU} retains PSA's access mechanism but populates the Historical State Index with historical state nodes assembled by predefined rules at subtask boundaries.

\paragraph{ISU improves task performance.}
Removing ISU reduces SR to zero in both environments, with lower PS as well. With PSA still available, this comparison suggests that states constructed and updated during execution better support task completion than rule-generated historical state nodes at subtask boundaries.

\paragraph{PSA supports task success with lower token usage.}
Compared with \emph{w/o PSA}, FlowState uses fewer tokens in both environments and improves all performance metrics in Group Travel Planning. Although \emph{w/o PSA} achieves slightly higher PS on Bundled Web Shopping, its SR remains below that of the full model. These results suggest that retaining all historical states does not necessarily improve end-to-end task success. By progressively disclosing relevant states and their supporting evidence as needed, PSA limits irrelevant history in the Active State, balancing task performance with token efficiency.

Overall, on the long-horizon, interdependent multi-session tasks evaluated in MemoryArena, ISU's continuous state construction and PSA's on-demand historical access play complementary roles. The full model shows strong task performance in both environments while using fewer tokens than the variant that exposes all historical states.

\begin{figure*}[!t]
\centering
\includegraphics[width=\textwidth]{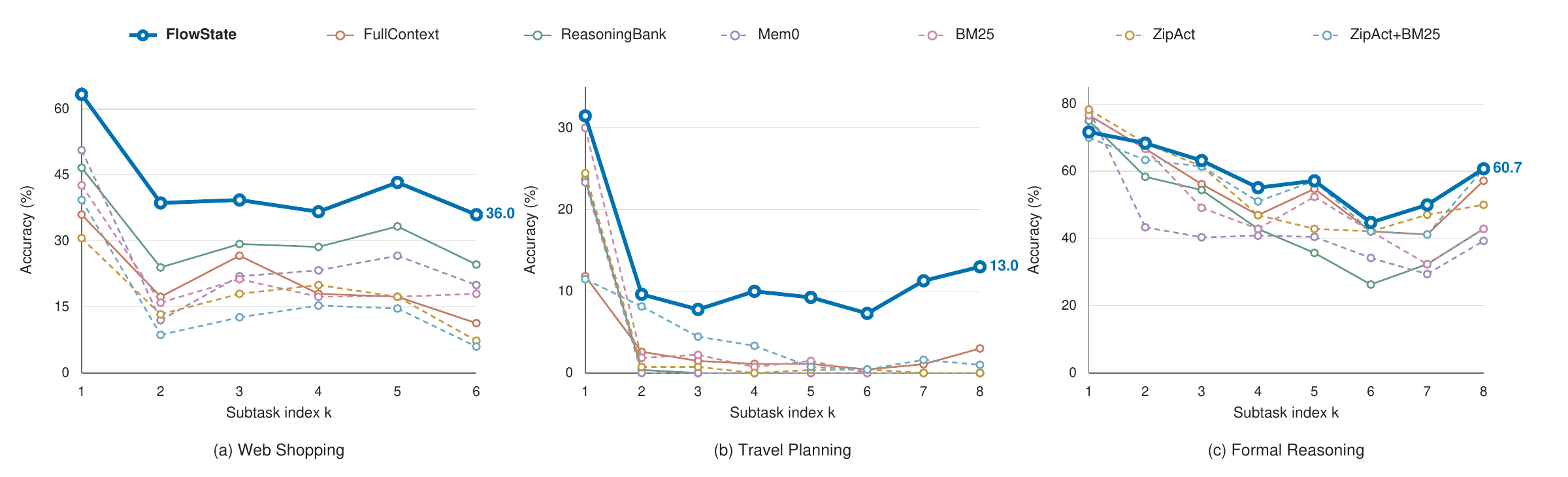}
 \caption{
        \textbf{Performance Across Subtask Depths.}
        Accuracy on (a) Web Shopping, (b) Travel Planning, and
        (c) Formal Reasoning.
        The horizontal axis denotes the subtask index $k$. The vertical axis reports
        $\mathrm{Acc}_k$, the percentage of task instances with at least $k$
        subtasks whose $k$-th subtask is completed correctly.
        Curves represent methods.
        The blue curve denotes FlowState, with annotations indicating
        its accuracy at the final reported depth in each panel.
        Vertical scales differ across panels.
    }
    \label{fig:accuracy-across-subtasks}
\end{figure*}

\subsection{FlowState Sustains Long-Horizon Advantages}

By expanding the dependencies and supporting evidence of historical states on demand, PSA enables the agent to continue using earlier information within a limited active context. Figure~\ref{fig:accuracy-across-subtasks} shows that FlowState maintains the highest accuracy at every reported subtask depth on Web Shopping and Travel Planning. At later depths on Travel Planning in particular, most baselines approach zero accuracy, while FlowState retains a clear advantage. On Formal Reasoning, FlowState generally leads at later depths and matches ZipAct+BM25 at the final depth.

Table~\ref{tab:relative-performance-retention} measures relative performance retention using $R_k=100\times\mathrm{Acc}_k/\mathrm{Acc}_1$. nAUC is the trapezoidal integral of the normalized trajectory divided by the depth range $K-1$, with higher values indicating better relative performance retention. Decline rate (DR) is the negative least-squares slope of the trajectory with respect to $k$, with lower values indicating a slower decline. FlowState achieves the highest nAUC in all three domains and the lowest DR on Travel Planning and Formal Reasoning. On Web Shopping, its DR is close to that of ReasoningBank, while its nAUC is higher.

These results show that FlowState combines high absolute accuracy on later subtasks with strong overall relative performance retention, consistent with PSA's design motivation of recovering earlier information on demand through structured access paths to support subsequent decisions.

\begin{table*}[t]
    \centering
    \caption{
        Relative performance retention across subtask depths.
        Metrics use accuracy normalized to each method's initial value.
        nAUC measures normalized curve area; DR is the fitted decline
        rate (percentage points per subtask).
        Best results are bolded and second-best results underlined.
    }
    \label{tab:relative-performance-retention}
    \small
    \renewcommand{\arraystretch}{1.18}
    \setlength{\tabcolsep}{0pt}
    \begin{tabular*}{0.85\textwidth}{
        @{\extracolsep{\fill}} lcccccc @{}
    }
        \toprule
        & \multicolumn{2}{c}{Web Shopping}
        & \multicolumn{2}{c}{Travel Planning}
        & \multicolumn{2}{c}{Formal Reasoning} \\
        \cmidrule(lr){2-3}
        \cmidrule(lr){4-5}
        \cmidrule(lr){6-7}
        Method
        & nAUC $\uparrow$ & DR $\downarrow$
        & nAUC $\uparrow$ & DR $\downarrow$
        & nAUC $\uparrow$ & DR $\downarrow$ \\
        \midrule
        Full Context
        & 57.22 & 10.48
        & 18.35 & \underline{7.30}
        & 69.82 & 4.63 \\
        ReasoningBank
        & \underline{64.71} & \textbf{5.06}
        & 7.37 & 8.43
        & 58.84 & 7.08 \\
        Mem0
        & 47.11 & 6.09
        & 7.14 & 8.33
        & 52.41 & 5.50 \\
        BM25
        & 47.97 & 8.26
        & 10.14 & 8.94
        & 64.33 & 6.52 \\
        ZipAct
        & 57.17 & 9.57
        & 8.48 & 8.54
        & 68.00 & 5.57 \\
        ZipAct+BM25
        & 37.63 & 10.61
        & \underline{31.04} & 12.51
        & \underline{77.87} & \underline{3.87} \\
        \midrule
        \textbf{FlowState}
        & \textbf{65.58} & \underline{5.65}
        & \textbf{35.17} & \textbf{4.66}
        & \textbf{80.66} & \textbf{3.68} \\
        \bottomrule
    \end{tabular*}
\end{table*}

\section{Conclusion}
\label{sec:conclusion}
We introduce FlowState, which turns execution states into continually updated, revisitable memory through Incremental State Update (ISU) and Progressive State Access (PSA). Experiments on MemoryArena and $\tau^3$-Bench show that FlowState substantially improves task performance while reducing token consumption, demonstrating the value of unifying state maintenance and historical access for long-horizon decision-making.

\subsection*{AI use statement}

We used generative AI tools to assist with checking relevant literature, writing research code, creating figures, and polishing the manuscript. The research questions, methodology, experimental design, and interpretation of results were developed by the authors. The authors take full responsibility for the final paper.

\subsection*{Ethics statement}

Our experiments use benchmark environments and involve no real-user data, human participants or annotators, or real-world transactions. Nevertheless, a system that retains information across interactions could expose sensitive information or carry outdated or incorrect assumptions into later decisions when deployed with real users. Our benchmark evaluation does not establish safety for deployment. Real-world applications would require controls over data access and retention, checks on the provenance and validity of information, and safeguards for consequential actions.

\subsection*{Reproducibility statement}
Section~\ref{sec:method} describes the state representation and the procedures for updating and accessing state. Section~\ref{sec:experiment-setup} specifies the benchmark domains, baselines, and evaluation metrics. Appendix~\ref{app:execution-rules} details the state transition rules, Appendix~\ref{app:runtime-prompts} presents the runtime prompt templates, and Appendix~\ref{app:experimental-details} reports model settings, benchmark revisions, prompt construction, and baseline implementations. We will release the complete source code at an appropriate time.

{
    \small
    \bibliography{iclr2027_conference}

@article{yao2022react,
  title={React: Synergizing reasoning and acting in language models},
  author={Yao, Shunyu and Zhao, Jeffrey and Yu, Dian and Du, Nan and Shafran, Izhak and Narasimhan, Karthik and Cao, Yuan},
  journal={arXiv preprint arXiv:2210.03629},
  year={2022}
}

@article{xu2026mem,
  title={A-mem: Agentic memory for llm agents},
  author={Xu, Wujiang and Liang, Zujie and Mei, Kai and Gao, Hang and Tan, Juntao and Zhang, Yongfeng},
  journal={Advances in Neural Information Processing Systems},
  volume={38},
  pages={17577--17604},
  year={2026}
}

@inproceedings{ouyang2026reasoningbank,
  title={Reasoningbank: Scaling agent self-evolving with reasoning memory},
  author={Ouyang, Siru and Yan, Jun and Hsu, I and Chen, Yanfei and Jiang, Ke and Wang, Zifeng and Han, Rujun and Le, Long and Daruki, Samira and Tang, Xiangru and others},
  booktitle={International Conference on Learning Representations},
  volume={2026},
  pages={94327--94354},
  year={2026}
}

@inproceedings{park2023generative,
  title={Generative agents: Interactive simulacra of human behavior},
  author={Park, Joon Sung and O'Brien, Joseph and Cai, Carrie Jun and Morris, Meredith Ringel and Liang, Percy and Bernstein, Michael S},
  booktitle={Proceedings of the 36th annual acm symposium on user interface software and technology},
  pages={1--22},
  year={2023}
}

@article{barres2025tau,
  title={{$\tau^2$-Bench}: Evaluating Conversational Agents in a Dual-Control Environment},
  author={Barres, Victor and Dong, Honghua and Ray, Soham and Si, Xujie and Narasimhan, Karthik},
  journal={arXiv preprint arXiv:2506.07982},
  year={2025}
}

@article{robertson2009bm25,
  title={The Probabilistic Relevance Framework: {BM25} and Beyond},
  author={Robertson, Stephen and Zaragoza, Hugo},
  journal={Foundations and Trends in Information Retrieval},
  year={2009},
  doi={10.1561/1500000019}
}

@article{deepseek2026v4,
  title={{DeepSeek-V4}: Towards Highly Efficient Million-Token Context Intelligence},
  author={{DeepSeek-AI} and others},
  journal={arXiv preprint arXiv:2606.19348},
  year={2026}
}

@misc{openai2026gpt56,
  author={{OpenAI}},
  title={{GPT-5.6} System Card},
  year={2026},
  howpublished={OpenAI Deployment Safety Hub},
  url={https://deploymentsafety.openai.com/gpt-5-6}
}

@misc{qwenteam2026qwen35,
  author={{Qwen Team}},
  title={{Qwen3.5}: Towards Native Multimodal Agents},
  year={2026},
  howpublished={Qwen Blog},
  url={https://qwen.ai/blog?id=qwen3.5}
}

@article{he2026memoryarena,
  title={Memoryarena: Benchmarking agent memory in interdependent multi-session agentic tasks},
  author={He, Zexue and Wang, Yu and Zhi, Churan and Hu, Yuanzhe and Chen, Tzu-Ping and Yin, Lang and Chen, Ze and Wu, Tong Arthur and Ouyang, Siru and Wang, Zihan and others},
  journal={arXiv preprint arXiv:2602.16313},
  year={2026}
}

@inproceedings{rozanov2025stateact,
  title={Stateact: Enhancing llm base agents via self-prompting and state-tracking},
  author={Rozanov, Nikolai and Rei, Marek},
  booktitle={Proceedings of the 1st Workshop for Research on Agent Language Models (REALM 2025)},
  pages={367--385},
  year={2025}
}

@article{singh2026agent,
  title={Agent-brace: Decoupling beliefs from actions in long-horizon tasks via verbalized state uncertainty},
  author={Singh, Joykirat and Khan, Zaid and Prasad, Archiki and Chen, Justin Chih-Yao and Nambi, Akshay and Lee, Hyunji and Stengel-Eskin, Elias and Bansal, Mohit},
  journal={arXiv preprint arXiv:2605.11436},
  year={2026}
}

@article{uddin2026ledgeragent,
  title={LedgerAgent: Structured State for Policy-Adherent Tool-Calling Agents},
  author={Uddin, Md Nayem and Saeidi, Amir and Blanco, Eduardo and Baral, Chitta},
  journal={arXiv preprint arXiv:2606.20529},
  year={2026}
}

@article{guo2026state,
  title={From State to Action: OODA-Tool for Reliable Multi-Turn Tool Use},
  author={Guo, Rongfeng and Huang, Yinxuan and Wu, Yusen and Zhong, Maoqing and Chen, Yunlu and Tang, Meng and Long, Teng and Hu, Vincent Tao},
  journal={arXiv preprint arXiv:2608.24368},
  year={2026}
}

@article{packer2023memgpt,
  title={Memgpt: Towards llms as operating systems},
  author={Packer, Charles and Wooders, Sarah and Lin, Kevin and Fang, Vivian and Patil, Shishir G and Stoica, Ion and Gonzalez, Joseph E},
  journal={arXiv preprint arXiv:2310.08560},
  year={2023}
}

@article{chhikara2025mem0,
  title={Mem0: Building production-ready ai agents with scalable long-term memory},
  author={Chhikara, Prateek and Khant, Dev and Aryan, Saket and Singh, Taranjeet and Yadav, Deshraj},
  journal={arXiv preprint arXiv:2504.19413},
  year={2025}
}

@article{rasmussen2025zep,
  title={Zep: a temporal knowledge graph architecture for agent memory},
  author={Rasmussen, Preston and Paliychuk, Pavlo and Beauvais, Travis and Ryan, Jack and Chalef, Daniel},
  journal={arXiv preprint arXiv:2501.13956},
  year={2025}
}

@article{lee2026minteval,
  title={MINTEval: Evaluating Memory under Multi-Target Interference in Long-Horizon Agent Systems},
  author={Lee, Hyunji and Chen, Justin Chih-Yao and Singh, Joykirat and Khan, Zaid and Stengel-Eskin, Elias and Bansal, Mohit},
  journal={arXiv preprint arXiv:2605.18565},
  year={2026}
}

@article{chao2026stale,
  title={STALE: Can LLM Agents Know When Their Memories Are No Longer Valid?},
  author={Chao, Hanxiang and Bai, Yihan and Sheng, Rui and Li, Tianle and Sun, Yushi},
  journal={arXiv preprint arXiv:2605.06527},
  year={2026}
}

@inproceedings{wang2026samem,
  title={SAMem: State-Aware Memory as a Fine-Grained Memory for LLM Agents in Decision-Making},
  author={Wang, Tong and Xu, Pei and Cao, Shiyue and Yang, Likun and Li, Daipeng and Jiao, Jianbin and Huang, Kaiqi},
  booktitle={Findings of the Association for Computational Linguistics: ACL 2026},
  pages={14691--14710},
  year={2026}
}

@article{ji2026memory,
  title={Memory is reconstructed, not retrieved: Graph memory for llm agents},
  author={Ji, Shuo and Li, Yibo and Hooi, Bryan},
  journal={arXiv preprint arXiv:2606.06036},
  year={2026}
}

@article{kaelbling1998planning,
  title={Planning and acting in partially observable stochastic domains},
  author={Kaelbling, Leslie Pack and Littman, Michael L and Cassandra, Anthony R},
  journal={Artificial intelligence},
  volume={101},
  number={1-2},
  pages={99--134},
  year={1998},
  publisher={Elsevier}
}

@article{liao2021dialogue,
  title={Dialogue state tracking with incremental reasoning},
  author={Liao, Lizi and Long, Le Hong and Ma, Yunshan and Lei, Wenqiang and Chua, Tat-Seng},
  journal={Transactions of the Association for Computational Linguistics},
  volume={9},
  pages={557--569},
  year={2021},
  publisher={MIT Press One Rogers Street, Cambridge, MA 02142-1209, USA journals-info~…}
}

@article{shinn2023reflexion,
  title={Reflexion: Language agents with verbal reinforcement learning},
  author={Shinn, Noah and Cassano, Federico and Gopinath, Ashwin and Narasimhan, Karthik and Yao, Shunyu},
  journal={Advances in neural information processing systems},
  volume={36},
  pages={8634--8652},
  year={2023}
}

@article{chen2026beyond,
  title={Beyond Semantic Organization: Memory as Execution State Management for Long-Horizon Agents},
  author={Chen, Yaoqi and Lai, Haibin and Feng, Yuru and Han, Chuyu and Zhang, Qianxi and Lu, Baotong and Li, Menghao and Wang, Xinjiang and Wang, Zhirui and Xu, Shusen and others},
  journal={arXiv preprint arXiv:2606.06090},
  year={2026}
}

@article{panzipact,
  title={ZipAct: Zipping Interaction History into a Compact State for Efficient LLM Agents},
  author={Pan, Zhiming and Luo, Junyu and Xiao, Zhiping and Ding, Kaize and Luo, Xiao and Zhang, Ming},
  journal={Transactions on Machine Learning Research},
  year    = {2026}
}

@article{ye2025agentfold,
  title={Agentfold: Long-horizon web agents with proactive context management},
  author={Ye, Rui and Zhang, Zhongwang and Li, Kuan and Yin, Huifeng and Tao, Zhengwei and Zhao, Yida and Su, Liangcai and Zhang, Liwen and Qiao, Zile and Wang, Xinyu and others},
  journal={arXiv preprint arXiv:2510.24699},
  year={2025}
}

@article{kang2025acon,
  title={Acon: Optimizing context compression for long-horizon llm agents},
  author={Kang, Minki and Chen, Wei-Ning and Han, Dongge and Inan, Huseyin A and Wutschitz, Lukas and Chen, Yanzhi and Sim, Robert and Rajmohan, Saravan},
  journal={arXiv preprint arXiv:2510.00615},
  year={2025}
}

@inproceedings{suzgun2026dynamic,
  title={Dynamic cheatsheet: Test-time learning with adaptive memory},
  author={Suzgun, Mirac and Yuksekgonul, Mert and Bianchi, Federico and Jurafsky, Dan and Zou, James},
  booktitle={Proceedings of the 19th Conference of the European Chapter of the Association for Computational Linguistics (Volume 1: Long Papers)},
  pages={7080--7106},
  year={2026}
}

@inproceedings{zhang2026agentic,
  title={Agentic context engineering: Evolving contexts for self-improving language models},
  author={Zhang, Qizheng and Hu, Changran and Upasani, Shubhangi and Ma, Boyuan and Hong, Fenglu and Kamanuru, Vamsidhar and Rainton, Jay and Wu, Chen and Ji, Mengmeng and Li, Hanchen and others},
  booktitle={International Conference on Learning Representations},
  volume={2026},
  pages={86069--86100},
  year={2026}
}
    \bibliographystyle{iclr2027_conference}
}

\clearpage

\appendix
\section{Related Work}
\label{sec:related_work}

\subsection{Execution State Tracking for Long-Horizon Agents}
In partially observable sequential decision-making, a belief state summarizes the action--observation history into the information needed for current decisions and is recursively updated based on the action taken and the new observation~\citep{kaelbling1998planning}. In task-oriented dialogue, Dialogue State Tracking updates user goals and relevant slot information as the conversation progresses. The incremental reasoning method of Liao et al. further combines the existing state with information from the current turn to progressively revise the estimate of the user's goals~\citep{liao2021dialogue}. Recent LLM agent methods explicitly maintain a task state or belief state that evolves during execution, organizing accumulated facts, constraints, and environmental information into a state representation for decision-making~\citep{rozanov2025stateact,singh2026agent,uddin2026ledgeragent,guo2026state}. ZipAct compresses the growing interaction history into a structured state comprising a Goal State, World State, and Constraint State, and decouples state updating from action generation, allowing the agent to act on the current state rather than the full history~\citep{panzipact}. These methods establish the value of maintaining a compact execution state. In long-horizon tasks, however, a further question is how an agent can revisit earlier states and their supporting evidence when a shift in task focus makes them relevant again.

\subsection{Long-Term Memory for LLM Agents}
Long-term memory methods store historical information in external memory so that information beyond the current context remains accessible in later interactions. Early work explored the preservation and reuse of historical information by accumulating observations and experiences, generating reusable reflections, or managing conversation history through tiered memory~\citep{park2023generative,shinn2023reflexion,packer2023memgpt}. More recent methods improve the organization and updating of historical information through memory consolidation, dynamic linking, graph structures, and temporal relations~\citep{chhikara2025mem0,xu2026mem,rasmussen2025zep}. In extended interactions involving multiple related tasks, agents must also use earlier actions and environmental feedback to guide later decisions. MemoryArena evaluates this need for memory in interdependent multi-session agentic tasks~\citep{he2026memoryarena}. As historical information accumulates and changes, its use presents further challenges: MINTEval reveals multi-target interference under frequent updates~\citep{lee2026minteval}, while STALE shows that, even when updated information is available, agents may fail to recognize that an older memory is no longer valid and adjust their behavior accordingly~\citep{chao2026stale}. Together, these works highlight the importance of not only
retaining historical information, but also selecting relevant memories and assessing their continued applicability as tasks and environments evolve.

\subsection{Connecting Execution State and Memory Access}

Recent work explores how accumulated experience can support
an agent's ongoing decisions.
Dynamic Cheatsheet maintains an evolving memory of strategies
and problem-solving insights for reuse at inference
time~\citep{suzgun2026dynamic}.
Agentic Context Engineering (ACE) develops this direction by
organizing context as an evolving playbook, accumulating and
refining strategies through generation, reflection, and
curation~\citep{zhang2026agentic}.
Other approaches tie memory access more directly to the current
decision context: SAMem retrieves state-aligned experiential
memories~\citep{wang2026samem}, while MRAgent uses intermediate
evidence to iteratively expand or prune access paths in an
associative memory graph~\citep{ji2026memory}.
Together, these methods connect past experience with current
decisions through reusable guidance and context-sensitive
memory access.

Concurrent work, MAGE~\citep{chen2026beyond}, studies agent memory
from the perspective of execution-state management, organizing
action--observation trajectories into a hierarchical state tree,
constructing working context along the active path, and supporting
revision through branching.
FlowState instead unifies current execution and cross-request
memory through persistent, semantically typed state nodes.
Nodes are individually addressable and retain relations and
references to raw tool observations, separating persistent retention
from visibility in the Active State.
PSA discloses historical states and supporting evidence on demand,
while ISU creates and revises current-request states informed by
the disclosed information.
Together, these mechanisms enable the same state representation
to support current decisions and cross-request reuse.

\section{Supplementary Materials}

\subsection{Performance and Efficiency Gains Across Base Models}

As shown in Table~\ref{tab:flowstate_model_comparison}, FlowState consistently improves the performance and token efficiency of Qwen3.5-397B-A17B, DeepSeek-V4-Flash, and GPT-5.6-terra. Across all nine model--task combinations, it improves PS and reduces token consumption, while increasing SR in eight cases and preserving it in the remaining one.

The gains are particularly pronounced on Web Shopping, where all three models improve PS by 17.55--21.78 percentage points and achieve nonzero SR from zero-success baselines. These performance gains are accompanied by substantial token savings in several settings. GPT-5.6-terra reduces token consumption by 84.30\% on Web Shopping, while DeepSeek-V4-Flash reduces token consumption by 73.01\% on Math and improves SR by 7.50 percentage points. Improvements also occur when baseline performance is already relatively high. On Physics, DeepSeek-V4-Flash increases PS from 72.50\% to 73.96\% while reducing token consumption by 51.69\%.

These results show that FlowState delivers consistent benefits across the evaluated models and tasks, combining improved task performance with lower token consumption even when baseline performance is already relatively high.

\definecolor{FSModelBg}{HTML}{EAF1FC}
\definecolor{FSDeltaBg}{HTML}{E8F3EC}
\definecolor{FSGreen}{HTML}{246B45}

\newcommand{\FSgain}[1]{%
  {\color{FSGreen}#1}%
}

\newcommand{\FStoken}[1]{%
  {\color{FSGreen}\bfseries #1}%
}

\begin{table}[t]
\centering
\caption{
FlowState improves task performance and reduces token usage
across three base models.
}
\label{tab:flowstate_model_comparison}

\small
\setlength{\tabcolsep}{4pt}
\renewcommand{\arraystretch}{1.16}

\resizebox{\textwidth}{!}{%
\begin{tabular}{l *{9}{r}}
\toprule
& \multicolumn{3}{c}{\textbf{Web Shopping}}
& \multicolumn{3}{c}{\textbf{Math}}
& \multicolumn{3}{c}{\textbf{Physics}} \\
\cmidrule(lr){2-4}
\cmidrule(lr){5-7}
\cmidrule(lr){8-10}
\textbf{Method}
& SR $\uparrow$ & PS $\uparrow$ & Tok. (M) $\downarrow$
& SR $\uparrow$ & PS $\uparrow$ & Tok. (M) $\downarrow$
& SR $\uparrow$ & PS $\uparrow$ & Tok. (M) $\downarrow$ \\
\midrule

\rowcolor{FSModelBg}
\multicolumn{10}{l}{\textbf{Qwen3.5-397B-A17B}} \\

Full Context
& 0.00 & 19.56 & 134.82
& 30.00 & 35.27 & 19.97
& 50.00 & 55.56 & 2.02 \\

FlowState
& 3.33 & 37.22 & 57.68
& 37.50 & 53.96 & 14.53
& 60.00 & 59.72 & 2.01 \\

\rowcolor{FSDeltaBg}
$\Delta$ (pp / \%)
& \FSgain{+3.33} & \FSgain{+17.66} & \FStoken{-57.22\%}
& \FSgain{+7.50} & \FSgain{+18.69} & \FStoken{-27.24\%}
& \FSgain{+10.00} & \FSgain{+4.16} & \FStoken{-0.50\%} \\

\addlinespace[5pt]

\rowcolor{FSModelBg}
\multicolumn{10}{l}{\textbf{DeepSeek-V4-Flash}} \\

Full Context
& 0.00 & 21.11 & 52.65
& 30.00 & 50.36 & 40.60
& 55.00 & 72.50 & 3.68 \\

FlowState
& 5.33 & 42.89 & 44.83
& 37.50 & 55.53 & 10.96
& 60.00 & 73.96 & 1.78 \\

\rowcolor{FSDeltaBg}
$\Delta$ (pp / \%)
& \FSgain{+5.33} & \FSgain{+21.78} & \FStoken{-14.86\%}
& \FSgain{+7.50} & \FSgain{+5.17}  & \FStoken{-73.01\%}
& \FSgain{+5.00} & \FSgain{+1.46}  & \FStoken{-51.69\%} \\

\addlinespace[5pt]

\rowcolor{FSModelBg}
\multicolumn{10}{l}{\textbf{GPT-5.6-terra}} \\

Full Context
& 0.00 & 33.89 & 313.95
& 42.50 & 54.23 & 14.59
& 50.00 & 62.60 & 2.76 \\

FlowState
& 4.00 & 51.44 & 49.29
& 45.00 & 59.25 & 8.28
& 50.00 & 67.65 & 1.62 \\

\rowcolor{FSDeltaBg}
$\Delta$ (pp / \%)
& \FSgain{+4.00} & \FSgain{+17.55} & \FStoken{-84.30\%}
& \FSgain{+2.50} & \FSgain{+5.02}  & \FStoken{-43.26\%}
& \FSgain{0.00}  & \FSgain{+5.05}  & \FStoken{-41.33\%} \\

\bottomrule
\end{tabular}%
}

\vspace{4pt}
\begin{minipage}{\textwidth}
\footnotesize
\textit{Note.}
SR and PS are percentages; token counts are in millions (M).
$\Delta$ compares FlowState with Full Context:
SR/PS changes are in percentage points (pp), while token changes
are relative percentages calculated from unrounded counts.
Negative token changes indicate savings.
\end{minipage}
\end{table}

\subsection{State Types and Semantic Scope}
\label{app:state-types}

Table~\ref{tab:state-types} summarizes the five state types and their semantic scope. These types organize information by its source and role in execution, distinguishing user requirements, environmental observations, and agent-generated assessments and outputs. The same taxonomy applies to both newly created states and historical states accessed during subsequent requests.

\begin{table}[!t]
\centering
\caption{State types and their semantic scope.}
\label{tab:state-types}
\small
\renewcommand{\arraystretch}{1.12}
\begin{tabularx}{\linewidth}{p{0.16\linewidth}p{0.17\linewidth}X}
\toprule
Perspective & State Type & Semantic Scope \\
\midrule
User
& preferences
& User preferences, constraints, and requirements extracted from requests. \\
Environment
& knowledge
& Task-relevant observations recorded from tool outputs. \\
Environment
& attributes
& Entity-centric information distilled from tool outputs, including properties and characteristics. \\
Agent
& judgements
& Agent assessments of task goals, plans, intermediate decisions, and execution risks. \\
Agent
& artifacts
& Reusable artifacts, tools, code, and other outputs created by the agent during task execution. \\
\bottomrule
\end{tabularx}
\end{table}

\subsection{Runtime Prompt Templates}
\label{app:runtime-prompts}

The templates below show the task-agent prompt structure for MemoryArena and
$\tau^3$-Bench. Each method has separate system and user messages. Braced fields
are populated at runtime; ReAct rules are shared, while policy and state rules
vary by task and state algorithm, respectively. Output protocols place
method-specific state fields before \texttt{tool\_calls} and
\texttt{assistant\_message}.
Ellipses in the FlowState status denote additional query or state entries.
Colored left rules mark the FlowState-specific instructions, output protocol, and state input.

\begingroup
\definecolor{flowpromptaccent}{RGB}{154,198,139}
\newcommand{\PromptLine}[1]{{\ttfamily\detokenize{#1}}\par}
\newcommand{\PromptFlowText}[1]{%
  \noindent\hbox{%
    {\color{flowpromptaccent}\vrule width 1.5pt}%
    \hspace{6pt}%
    \parbox[t]{\dimexpr\linewidth-7.5pt\relax}{%
      \raggedright\ttfamily #1}}\par\vspace{1pt}}
\newcommand{\PromptFlowLine}[1]{\PromptFlowText{\detokenize{#1}}}
\newcommand{\PromptFlowIndentLine}[2]{%
  \PromptFlowText{\setlength{\leftskip}{#1em}\detokenize{#2}}}
\newcommand{\PromptFlowHeading}[2]{%
  \PromptFlowText{\bfseries\char35{}\char35{}\ifnum#1=3\char35{}\fi\space\detokenize{#2}}%
  \nopagebreak[4]}
\newcommand{\PromptIndentLine}[2]{\noindent\hspace*{#1em}{\ttfamily\detokenize{#2}}\par}
\newcommand{\PromptBraceLine}[3]{\noindent\hspace*{#1em}{\ttfamily\char#2\detokenize{#3}}\par}
\newcommand{\PromptHeader}[1]{%
  \colorbox[gray]{0.93}{%
    \parbox{\dimexpr\linewidth-2\fboxsep\relax}{%
      \centering\bfseries\fontsize{8}{9}\selectfont #1}}\par}
\newcommand{\PromptCard}[3]{%
  \par\vspace{4pt}%
  \ifdim\dimexpr\pagegoal-\pagetotal\relax<72pt\newpage\fi
  \noindent\hrule\vspace{1pt}
  {\centering\bfseries\fontsize{9}{10}\selectfont #1\par}
  \vspace{1pt}
  \begingroup
  \fontsize{9}{10}\selectfont
  \setlength{\parindent}{0pt}
  \setlength{\parskip}{0pt}
  \PromptHeader{System Instruction}
  #2
  \vspace{1pt}
  \PromptHeader{Input Prompt}
  #3
  \par\endgroup
  \vspace{1pt}\hrule\par\vspace{4pt}}

\PromptCard{FlowState}{%
  \PromptLine{<instructions>}
  \PromptLine{{react_rules}}
  \PromptFlowHeading{2}{flowstate_rules}
  \PromptFlowLine{Use the Active State block in <agent_status> as your working memory. Before each action, check whether it contains the facts and identifiers required for that action.}
  \PromptFlowHeading{3}{Incremental State Update}
  \PromptFlowLine{- Update Active State whenever task-relevant facts, requirements, plans, or results change. Output only the changes; do not regenerate the entire state.}
  \PromptFlowLine{- Assign each state to preference (user requirements and constraints), knowledge (tool observations), attribute (entity properties), judgement (goals, plans, decisions, and risks), or artifacts (reusable outputs).}
  \PromptFlowLine{- Apply state_delta to the Active State block:}
  \PromptFlowIndentLine{1}{- add: Insert a new state with a new id.}
  \PromptFlowIndentLine{1}{- update: Revise an existing Active State entry by its id.}
  \PromptFlowIndentLine{1}{- remove: Delete an existing Active State entry by its id.}
  \PromptFlowLine{- Do not update or remove entries whose ids appear in the Historical State Index, even if their full contents have been expanded into Active State. To revise historical information, add a new Active State entry and link it to the historical state through refs.}
  \PromptFlowLine{- Give each state a semantic id, a concise summary, and a structured value. Link it to supporting evidence and related states through refs.}
    \PromptFlowLine{- Declare directed relations through refs. Every edge points from the state being added or updated (source) to the state_id in that ref (target). All relation types use this direction. Set rel to one of: supports, derives, follows, supersedes.}
  \PromptFlowHeading{3}{Progressive State Access}
  \PromptFlowLine{- If required information is missing, use prior queries and state summaries in the Historical State Index, together with related_states in Active State, to select relevant historical content. Use the corresponding state_id as state_focus.target_id.}
  \PromptFlowLine{- Specify what information you need and which decision or action it enables. Never infer exact facts or identifiers from summaries.}
  \PromptFlowLine{- A related_states entry identifies a related state but does not reveal its full content. Request that state separately when its details are needed.}
  \PromptFlowLine{- Use newly disclosed content only from the next model invocation onward. Defer any action that depends on it. Do not request content already visible.}
  \PromptLine{</instructions>}
  \PromptLine{<policy>}
  \PromptLine{{task_policy}}
  \PromptLine{</policy>}
  \PromptLine{<tools>}
  \PromptLine{{available_tools}}
  \PromptLine{</tools>}
  \PromptLine{<output_protocol>}
  \nopagebreak[4]
  \PromptFlowLine{Return exactly four blocks in the following order. Use valid JSON inside each block. Do not include code fences or text outside the blocks.}
  \nopagebreak[4]
  \PromptFlowLine{```text}
  \noindent\begin{minipage}{\linewidth}
  \PromptFlowLine{<state_focus>}
  \PromptFlowLine{[{"target_id": "<visible_id>", "reason": "<missing information and intended use>"}]}
  \PromptFlowLine{</state_focus>}
  \end{minipage}\par
  \noindent\begin{minipage}{\linewidth}
  \PromptFlowLine{<state_delta>}
  \PromptFlowLine{[{"state_set": "<state_type>", "operation": "add", "state_id": "<state_id>", "summary": "<concise summary>", "value": {}, "refs": [{"rel": "<relation>", "state_id": "<source_state_id>"}]}]}
  \PromptFlowLine{</state_delta>}
  \end{minipage}\par
  \noindent\begin{minipage}{\linewidth}
  \PromptFlowLine{<tool_calls>}
  \PromptFlowLine{[{"name": "<tool_name>", "arguments": {}}]}
  \PromptFlowLine{</tool_calls>}
  \end{minipage}\par
  \noindent\begin{minipage}{\linewidth}
  \PromptFlowLine{<assistant_message>}
  \PromptFlowLine{{"content": null}}
  \PromptFlowLine{</assistant_message>}
  \PromptFlowLine{```}
  \end{minipage}\par
  \PromptFlowLine{Set operation to add, update, or remove. For an execution decision, request a tool call or provide a user-facing message.}
  \PromptFlowLine{At each step, jointly propose state updates, historical access, and an execution decision whenever the available information supports them. Do not split independent operations across steps.}
  \PromptFlowLine{Request missing historical information and perform any independent operations in the same step. Wait for the requested information before making updates or taking actions that require it.}
  \PromptFlowLine{Always output all four protocol blocks. Use [] for unused arrays and set assistant_message.content to null when no user-facing reply is needed. At least one operation must be non-empty.}
  \PromptLine{</output_protocol>}
}{%
  \begingroup
  \renewcommand{\PromptLine}[1]{\PromptFlowLine{#1}}
  \renewcommand{\PromptIndentLine}[2]{\PromptFlowText{\hspace*{#1em}\detokenize{#2}}}
  \renewcommand{\PromptBraceLine}[3]{\PromptFlowText{\hspace*{#1em}\char#2\detokenize{#3}}}
  \PromptLine{<agent_status>}
  \PromptLine{Historical State Index:}
  \PromptLine{[}
  \PromptBraceLine{1}{123}{}
  \PromptIndentLine{2}{"user_query": "{prior_query}",}
  \PromptIndentLine{2}{"query_id": {prior_query_id},}
  \PromptIndentLine{2}{"states": [}
  \PromptBraceLine{3}{123}{}
  \PromptIndentLine{4}{"state_id": "{historical_state_id}",}
  \PromptIndentLine{4}{"summary": "{historical_state_summary}"}
  \PromptBraceLine{3}{125}{}
  \PromptIndentLine{2}{]}
  \PromptBraceLine{1}{125}{,}
  \PromptIndentLine{1}{...}
  \PromptLine{]}
  \PromptLine{Active State:}
  \PromptLine{[}
  \PromptBraceLine{1}{123}{}
  \PromptIndentLine{2}{"state_id": "{active_state_id}",}
  \PromptIndentLine{2}{"type": "{state_type}",}
  \PromptIndentLine{2}{"related_states": [}
  \PromptIndentLine{3}{{"state_id":"{related_state_id}","rel":"{relation}"}}
  \PromptIndentLine{2}{],}
  \PromptIndentLine{2}{"summary": "{active_state_summary}",}
  \PromptIndentLine{2}{"value": {state_value_json}}
  \PromptBraceLine{1}{125}{,}
  \PromptIndentLine{1}{...}
  \PromptLine{]}
  \PromptLine{</agent_status>}
  \endgroup
  \PromptLine{<latest_observation>}
  \PromptLine{{observation}}
  \PromptLine{</latest_observation>}
  \PromptLine{<user_query>}
  \PromptLine{{query}}
  \PromptLine{</user_query>}
}

\PromptCard{Full Context}{%
  \PromptLine{<instructions>}
  \PromptLine{{react_rules}}
  \PromptLine{</instructions>}
  \PromptLine{<policy>}
  \PromptLine{{task_policy}}
  \PromptLine{</policy>}
  \PromptLine{<tools>}
  \PromptLine{{available_tools}}
  \PromptLine{</tools>}
  \PromptLine{<output_protocol>}
  \PromptLine{{react_output_protocol}}
  \PromptLine{</output_protocol>}
}{%
  \PromptLine{<trajectory_history>}
  \PromptLine{{full_trajectory}}
  \PromptLine{</trajectory_history>}
  \PromptLine{<latest_observation>}
  \PromptLine{{observation}}
  \PromptLine{</latest_observation>}
  \PromptLine{<user_query>}
  \PromptLine{{query}}
  \PromptLine{</user_query>}
}

\PromptCard{Memory System}{%
  \PromptLine{<instructions>}
  \PromptLine{{react_rules}}
  \PromptLine{</instructions>}
  \PromptLine{<policy>}
  \PromptLine{{task_policy}}
  \PromptLine{</policy>}
  \PromptLine{<tools>}
  \PromptLine{{available_tools}}
  \PromptLine{</tools>}
  \PromptLine{<output_protocol>}
  \PromptLine{{react_output_protocol}}
  \PromptLine{</output_protocol>}
}{%
  \PromptLine{<memory>}
  \PromptLine{{retrieved_memory}}
  \PromptLine{</memory>}
  \PromptLine{<trajectory_history>}
  \PromptLine{{query_trajectory}}
  \PromptLine{</trajectory_history>}
  \PromptLine{<latest_observation>}
  \PromptLine{{observation}}
  \PromptLine{</latest_observation>}
  \PromptLine{<user_query>}
  \PromptLine{{query}}
  \PromptLine{</user_query>}
}

\PromptCard{State Tracking}{%
  \PromptLine{<instructions>}
  \PromptLine{{react_rules}}
  \PromptLine{{state_tracking_rules}}
  \PromptLine{</instructions>}
  \PromptLine{<policy>}
  \PromptLine{{task_policy}}
  \PromptLine{</policy>}
  \PromptLine{<tools>}
  \PromptLine{{available_tools}}
  \PromptLine{</tools>}
  \PromptLine{<output_protocol>}
  \PromptLine{{state_output_protocol}}
  \PromptLine{</output_protocol>}
}{%
  \PromptLine{<agent_status>}
  \PromptLine{{execution_state}}
  \PromptLine{</agent_status>}
  \PromptLine{<latest_observation>}
  \PromptLine{{observation}}
  \PromptLine{</latest_observation>}
  \PromptLine{<user_query>}
  \PromptLine{{query}}
  \PromptLine{</user_query>}
}

\PromptCard{State Tracking + Memory}{%
  \PromptLine{<instructions>}
  \PromptLine{{react_rules}}
  \PromptLine{{state_tracking_rules}}
  \PromptLine{</instructions>}
  \PromptLine{<policy>}
  \PromptLine{{task_policy}}
  \PromptLine{</policy>}
  \PromptLine{<tools>}
  \PromptLine{{available_tools}}
  \PromptLine{</tools>}
  \PromptLine{<output_protocol>}
  \PromptLine{{state_output_protocol}}
  \PromptLine{</output_protocol>}
}{%
  \PromptLine{<memory>}
  \PromptLine{{retrieved_memory}}
  \PromptLine{</memory>}
  \PromptLine{<agent_status>}
  \PromptLine{{execution_state}}
  \PromptLine{</agent_status>}
  \PromptLine{<latest_observation>}
  \PromptLine{{observation}}
  \PromptLine{</latest_observation>}
  \PromptLine{<user_query>}
  \PromptLine{{query}}
  \PromptLine{</user_query>}
}

\endgroup

\subsection{Execution Rules}
\label{app:execution-rules}

For request $q_k$ at decision step $t$, the execution rules $P$ define the
admissibility conditions for the model-proposed state delta $\Delta S_{k,t}$
and access set $R_{k,t}$. The runtime applies these conditions through
$\operatorname{Validate}_{\mathrm{upd}}$ and
$\operatorname{Validate}_{\mathrm{acc}}$, yielding the validated requests
$\widehat{\Delta S}_{k,t}$ and $\widehat{R}_{k,t}$, respectively.

\paragraph{Incremental State Update.}
Each $\mathsf{Add}$ operation in $\Delta S_{k,t}$ requires an unused semantic
identifier. The target of each $\mathsf{Update}$ or $\mathsf{Remove}$ operation must be present in
$S_{k,t}$ and must have been created during the current request. References in proposed
state relations must resolve to valid node identifiers. Operations failing
these conditions are excluded from $\widehat{\Delta S}_{k,t}$.

\paragraph{Progressive State Access.}
Let $\operatorname{IDs}(G_k)$ denote the historical state identifiers in
$G_k$, and let $\operatorname{ToolRefs}(S_{k,t})$ denote the identifiers
of raw tool results referenced by states in the current Active State
$S_{k,t}$. The validated access set is restricted to
\begin{equation}
\widehat{R}_{k,t}\subseteq R_{k,t}\cap
\bigl(\operatorname{IDs}(G_k)\cup
\operatorname{ToolRefs}(S_{k,t})\bigr).
\label{eq:psa-admissible-targets}
\end{equation}
Requested identifiers outside this set are rejected by
$\operatorname{Validate}_{\mathrm{acc}}$ and excluded from
$\widehat{R}_{k,t}$.

\subsection{Experimental and Implementation Details}
\label{app:experimental-details}

\paragraph{Benchmarks and evaluation scope.}
MemoryArena tests continual reuse of historical information across interdependent subtasks.
We evaluate a subset of its environments: Bundled Web Shopping requires compatibility-aware purchases across sessions; Group Travel Planning introduces travelers and interdependent preferences over time; and Formal Reasoning (Math and Physics) requires reusing intermediate results in later derivations. In $\tau^3$-Bench, Airline and Retail involve policy-constrained interactions with simulated users, including information lookup, flight rebooking, and refunds.

We exclude Progressive Web Search because we identified issues with its ground-truth annotations that undermine the reliability of answer-based evaluation. To avoid confounding method comparisons with annotation errors, we omit this environment from all reported results and aggregate metrics.

The evaluated MemoryArena task sets comprise 150 shopping tasks (900 subtasks), 270 travel-planning tasks (1,869 subtasks), 40 math tasks (354 subtasks), and 20 physics tasks (86 subtasks). For $\tau^3$-Bench, we use the base task sets of 50 airline tasks and 114 retail tasks. The evaluation across base models uses the Web Shopping, Math, and Physics task sets, while the component ablation uses the Web Shopping and Travel Planning task sets. The analysis of performance across subtask depths reuses the main evaluation data and includes only tasks containing the corresponding subtask.

\paragraph{Evaluation metrics.}
On MemoryArena, SR requires a final bundle or plan satisfying all group members in Shopping and Travel, or a correct final subtask in Math and Physics. PS averages the fraction of correctly completed subtasks per task for Shopping, Math, and Physics; for Travel, it is the fraction of correctly completed subtasks among all evaluated subtasks. sPS credits satisfied constraints within each Travel subtask. On $\tau^3$-Bench, Pass Rate measures task success, DB Accuracy the correctness of the final database state, and Action Accuracy measures reference-action matching, not overall action correctness. \#Tok includes prompt and generation tokens.

\paragraph{Model configurations.}
We evaluate FlowState with three backbone models: DeepSeek-V4-Flash, GPT-5.6-terra, and Qwen3.5-397B-A17B. The full-context
baseline is evaluated with all three models, whereas the memory-system,
state-tracking, and hybrid baselines use DeepSeek-V4-Flash.
For agent inference, we specify the vLLM request parameters
\texttt{temperature=0.1}, \texttt{top\_p=1.0}, and
\texttt{max\_completion\_tokens=32768}, with thinking disabled
(\texttt{enable\_thinking=false}). No other generation parameters are
overridden.

\paragraph{Benchmark protocol and prompt construction.}
We use the MemoryArena and $\tau^3$-Bench implementations at the revisions
specified below.\footnote{MemoryArena revision:
\href{https://github.com/ZexueHe/MemoryArena/commit/6cd9de14b71915e39ac742a20dc33785e14b6aab}{\texttt{6cd9de1}};
$\tau^3$-Bench v1.0.1:
\href{https://github.com/sierra-research/tau2-bench/commit/fc0055dc4e0a316c3f83133267fbd6faaa770992}{\texttt{fc0055d}}.}
For the domains in Section~\ref{sec:experiment-setup}, we retain the benchmark
task configurations, environment tools and their descriptions, and evaluation
procedures. We adapt the task-agent prompt assembly to the system- and
user-message structures in Appendix~\ref{app:runtime-prompts}, combining
benchmark task policies and tool descriptions with method-specific instructions
and context.
All methods compared within a domain use the same benchmark tasks,
environment tools, and evaluation procedures.

\paragraph{Baselines.}
Full Context appends the available interaction history to the agent context without explicit state abstraction or retrieval.
ReasoningBank, BM25, and Mem0 store and retrieve historical information.
Our ReasoningBank and BM25 baselines follow the implementations distributed with MemoryArena;
we reproduce Mem0 from its publicly released code.\footnote{Mem0 revision:
\href{https://github.com/mem0ai/mem0/commit/8d6c001966573786d36908bfe4dfd52935749200}{\texttt{8d6c001}}.}
We retain the released memory-agent prompts for ReasoningBank and Mem0.
ZipAct maintains a compact execution state and is reproduced from its publicly released code.\footnote{ZipAct revision:
\href{https://github.com/Thomas-mci-21/ZipAct_TMLR/commit/f0258044c3be203d1e2edcb8d2559cbdf3c5de00}{\texttt{f025804}}.}
For ZipAct, we populate the domain-specific injectables in its state-construction
prompts with state items relevant to each benchmark task, as specified by
\citet{panzipact}; the remaining state-module prompt content follows the
released implementation.
ZipAct+BM25 augments ZipAct with BM25 retrieval over the interaction history.

\paragraph{Ablation Design and Scope.}
To evaluate the roles of ISU and PSA in the full framework, we construct each ablation variant by replacing one component with an alternative mechanism while preserving the functionality of the other.
Specifically, \textbf{w/o PSA} retains ISU and directly includes the full contents of all historical states in the Active State, comparing full exposure of historical states with on-demand access.
Conversely, \textbf{w/o ISU} retains PSA and constructs a state node from the final response at the end of each query, linking it to the raw tool results from that query through fixed rules without involving ISU.
This variant still supports on-demand access to historical final responses and their associated tool results.
Thus, the w/o ISU ablation compares incremental state maintenance during execution with final-response-based state construction at query boundaries, rather than the presence or absence of access to historical information or evidence.
These ablations assess the overall contribution of each mechanism without isolating the contributions of its individual design choices.

\end{document}